\documentclass[conference]{IEEEtran}
\IEEEoverridecommandlockouts
\usepackage{cite}
\usepackage{amsmath,amssymb,amsfonts}
\usepackage{algorithmic}
\usepackage{graphicx}
\usepackage{textcomp}
\usepackage{xcolor}
\usepackage{tikz}

\def\BibTeX{{\rm B\kern-.05em{\sc i\kern-.025em b}\kern-.08em
    T\kern-.1667em\lower.7ex\hbox{E}\kern-.125emX}}
\begin{document}

\title{Learning Using Privileged Information for Litter Detection
}

\author{\IEEEauthorblockN{Matthias Bartolo}
\IEEEauthorblockA{\textit{Dept. of Artificial Intelligence} \\
\textit{University of Malta}\\
Msida, Malta \\
matthias.bartolo@um.edu.mt}
\and
\IEEEauthorblockN{Konstantinos Makantasis}
\IEEEauthorblockA{\textit{Dept. of Artificial Intelligence} \\
\textit{University of Malta}\\
Msida, Malta \\
konstantinos.makantasis@um.edu.mt}
\and
\IEEEauthorblockN{Dylan Seychell}
\IEEEauthorblockA{\textit{Dept. of Artificial Intelligence} \\
\textit{University of Malta}\\
Msida, Malta \\
dylan.seychell@um.edu.mt}
}

\maketitle
\renewcommand{\thefootnote}{}
\footnotetext{This paper was accepted at the 13th European Workshop on Visual Information Processing (EUVIP 2025).}
\renewcommand{\thefootnote}{\arabic{footnote}}


\thispagestyle{plain}
\pagestyle{plain}

\begin{abstract}
As litter pollution continues to rise globally, developing automated tools capable of detecting litter effectively remains a significant challenge. This study presents a novel approach that combines, for the first time, privileged information with deep learning object detection to improve litter detection while maintaining model efficiency. We evaluate our method across five widely used object detection models, addressing challenges such as detecting small litter and objects partially obscured by grass or stones. In addition to this, a key contribution of our work can also be attributed to formulating a means of encoding bounding box information as a binary mask, which can be fed to the detection model to refine detection guidance.
Through experiments on both within-dataset evaluation on the renowned SODA dataset and cross-dataset evaluation on the BDW and UAVVaste litter detection datasets, we demonstrate consistent performance improvements across all models. Our approach not only bolsters detection accuracy within the training sets but also generalises well to other litter detection contexts. Crucially, these improvements are achieved without increasing model complexity or adding extra layers, ensuring computational efficiency and scalability. Our results suggest that this methodology offers a practical solution for litter detection, balancing accuracy and efficiency in real-world applications.
\end{abstract}

\begin{IEEEkeywords}
Litter Detection, Learning Using Privileged Information, Computer Vision, Knowledge Distillation, Object Detection
\end{IEEEkeywords}


\section{Introduction}
Litter pollution remains a stagnant issue, with ramifications that extend beyond environmental deterioration to encompass broader socio-economic instability. With global waste output projected to rise from 2.1 to 2.6 billion tonnes annually by 2030 \cite{kaza2018waste}, the limitations of current management systems are becoming increasingly apparent. In response to this global challenge, recent research \cite{zerowaste, taco2020} has begun to explore the application of Artificial Intelligence (AI), particularly computer vision techniques, as a means of automating the detection of litter in various environments. Similarly, Unmanned Aerial Vehicle (UAV) technology has received growing attention for its potential to assist in detecting litter across wide or inaccessible areas \cite{uavvaste, soda_dataset}.

However, despite recent progress, there still remains a clear need to improve the accuracy and efficiency of these technologies. Achieving optimal performance in diverse and dynamic environments continues to present significant challenges, especially in balancing detection accuracy with inference speed. In practical applications, litter frequently includes transparent materials or items that are either very small or partially concealed by natural elements such as grass or stones. These conditions necessitate more complex architectural frameworks and a more rigorous approach to model training, such as incorporating knowledge distillation techniques to improve generalisation while maintaining computational efficiency.
It is within this context that this paper proposes the following:
\begin{enumerate}
    \item A novel methodology that integrates privileged information and deep learning object detection models to improve litter detection, without increasing the number of model parameters or affecting inference time.
    \item A performance evaluation of this methodology across five widely-used object detectors.
    \item A detailed examination of the proposed methodology using the SODA dataset \cite{soda_dataset}, alongside cross-validation on the BDW \cite{bottle_detection} and UAVVaste \cite{uavvaste} litter detection datasets from aerial imagery.
\end{enumerate}

\section{Related Work}
Computer vision has gained attention in addressing environmental issues, especially waste detection. Litter detection stands out due to its relevance to sustainability and public hygiene, prompting the development of datasets and automated detection methods.

\subsection{Litter Detection}
In recent years, a number of litter detection datasets and methods have been introduced to support research in automated litter detection. Wang et al. \cite{bottle_detection} introduced the UAV-Bottle, or BDW, dataset in 2018, which includes 25,407 UAV-captured images focused solely on the detection of bottles across diverse environments. In addition to UAV-based litter detection, Proença and Simões \cite{taco2020} developed the TACO dataset in 2020. Comprising 1,500 images across 60 categories, this dataset broadened the scope of litter detection tasks and continues to be widely used in related research. In the same year, Wang et al. \cite{mju_waste} released the MJU-WASTE dataset, which provides 2,475 images dedicated to litter segmentation within a single waste category. Similarly, Kraft et al. \cite{uavvaste} introduced the UAVVaste dataset in 2021, focusing on UAV-based litter detection. This dataset contains 772 UAV images and addresses the challenges of detecting small objects within a single waste category. Additionally, in terms of non-UAV based litter detection, Bashkirov et al. \cite{zerowaste} developed the ZeroWaste dataset, while Córdova et al. \cite{plastopol} created the PlastOPol dataset, containing 4,503 and 2,418 images, respectively, providing real-world data that further improves litter detection research. Most recently, Pisani et al. \cite{soda_dataset, detect_litter} presented the SODA dataset in 2024, which includes 829 images captured at various UAV altitudes across six categories. 
Across all of these approaches, the authors utilised the curated datasets to develop effective litter detection models, employing methodologies similar to those used in object detection, which involve training prominent deep learning detection architectures. Notable detectors that were trained in the aforementioned approaches, include YOLO \cite{yolo}, Faster R-CNN \cite{fasterrcnn}, SSD \cite{ssd}, and RetinaNet \cite{retinanet}, among others. In addition, pre-processing techniques such as tiling and data augmentation were also commonly employed to bolster training robustness and accuracy \cite{detect_litter, uavvaste}. Nevertheless, in all of these approaches, the repeated trend of improving accuracy by exploring or developing complex architectures and learning paradigms necessitates a clearer way forward \cite{soda_dataset, detect_litter, uavvaste}.

\subsection{Learning Using Privileged Information in Computer Vision}
The Learning using Privileged Information (LUPI) paradigm, introduced by Vapnik and Vashist \cite{lupi, Vapnik2015LearningUP}, expands traditional learning tasks by incorporating supplementary data alongside the standard input/output training pairs in machine learning. This additional information is often more pertinent to the task at hand, thereby improving prediction accuracy. The concept of LUPI allows for the \textit{transfer of knowledge} from a teacher, trained with privileged data, to a student who only has access to the input information.
In the field of Computer Vision, several problems present an asymmetric distribution of information between training and test phases \cite{learning2rank}, making LUPI particularly applicable. Sharmanska et al. \cite{learning2rank} investigate four types of privileged information for object classification: semantic properties, bounding boxes, tags, and annotator rationale. Their study shows that applying LUPI to the SVM+ algorithm improves performance.
In a similar study, Wang et al. \cite{lupi_classification} address the same issue by applying similarity constraints to capture the relationship between available and privileged information. The authors use high-resolution images and image tags as privileged data, which are accessible during training but not during testing.

\subsection{Knowledge Distillation in Computer Vision}
Knowledge distillation is a pivotal technique in machine learning that allows the transfer of knowledge from a large, complex model to a smaller, more efficient one. In the context of computer vision, as discussed by \cite{distillation1}, there are various methods for achieving this, including response-based, feature-based, and relation-based knowledge transfer. These approaches can be applied across a wide range of vision tasks, such as image classification, object detection, and multimodal vision models \cite{distillation1}. Focusing on object detection, two common distillation techniques are feature imitation and logit mimicking \cite{distillation2}. Interestingly, the use of valuable localisation regions to selectively distil both classification and localisation knowledge for specific areas is another key aspect of this process, as proposed in \cite{distillation2}.

In summary, existing litter detection methods rely on complex models and large datasets to boost accuracy. In this context, applying privileged information during training without altering model structure or inference speed offers a promising alternative.

\begin{figure*}[ht]
    \centering
    \includegraphics[width=1\textwidth]{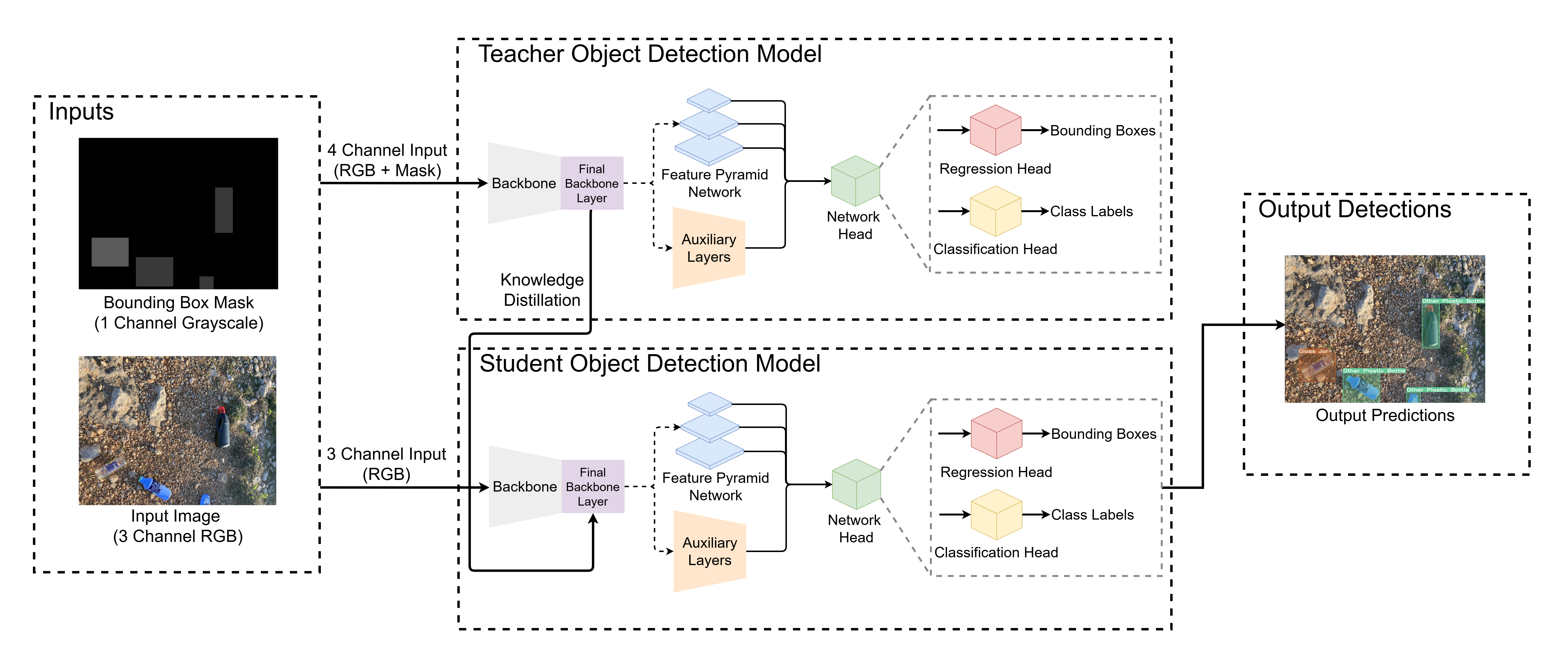}
    \caption{Architecture of the object detection models, illustrating the use of the LUPI paradigm with RGB and bounding box mask inputs, the teacher and student networks, the final backbone layer for knowledge distillation, and the output predictions.}
    \label{fig:architecture}
\end{figure*}

\section{Methodology}
This section presents our methodological framework. We begin by defining the problem and articulating our conceptual approach. Subsequently, we provide the implementation detains a description of the experimental protocol.
\subsection{Problem Definition}
This study proposes a novel methodology that applies learning with privileged information to the task of object detection, specifically focusing on litter detection. Although the LUPI paradigm has previously been explored within computer vision, particularly in relation to image classification, its application to object detection remains unexplored. In this regard, the object detection problem within the LUPI framework can be rigorously described as follows: consider a training set of triplets as defined in Equation \eqref{eq:object_detection}.
\begin{equation} \label{eq:object_detection}
\mathcal D = {(x_i, x_i^*, y_i)}_{i=1}^N, \quad x_i \in X , x_i^* \in X^*, y_i \in Y.
\end{equation}
In this formulation, $X$ represents the space of input images, $X^*$ denotes the space of privileged information instances, and $Y$ comprises the space of bounding boxes with their associated class labels. Given a teacher model defined as:
\begin{equation}
f_{teacher}: X \cup X^* \rightarrow Y,
\end{equation}
which accurately predicts $y$ based on both $x$ and $x^*$, our objective is to develop a student model:
\begin{equation}
f_{student}: X \rightarrow Y,
\end{equation}
that effectively maps $X$ to $Y$ by leveraging not only the intrinsic information in $X$, but also the knowledge encoded within $f_{teacher}$. In other words, during training, $f_{student}$ learns to map $X$ to $Y$ through knowledge distillation from $f_{teacher}$ and the information contained in the labeled examples from the training set $\mathcal D$.

\subsection{Our Approach}
In this study, both $f_{teacher}$ and $f_{student}$ are implemented as neural networks, each comprising $L$ layers. These models can be formally expressed as:
\begin{equation}
f_{teacher} = f_1^{(t)} \circ f_2^{(t)} \circ \cdots \circ f_l^{(t)} \circ \cdots \circ f_L^{(t)},
\end{equation}
\begin{equation}
f_{student} = f_1^{(s)} \circ f_2^{(s)} \circ \cdots \circ f_l^{(s)} \circ \cdots \circ f_L^{(s)},
\end{equation}
Here, “$\circ$” represents the function composition operation, while $f_i^{(t)}$ and $f_i^{(s)}$ denote the $i$-th layer of the teacher and student models, respectively. We establish the constraint that the $l$-th layer of both the teacher and student networks contain an identical number of hidden neurons. Consequently, knowledge can be distilled from the teacher to the student model by minimizing the dissimilarity:
\begin{equation}
D(f_l^{(t)}, f_l^{(s)}).
\end{equation}
This minimization is performed for each triplet $(x_i, x_i^*, y_i)$ in the training set $\mathcal D$. Specifically, given a triplet $(x_i, x_i^*, y_i)$, we require that the latent representation at the $l$-th layer of the student closely approximates the corresponding latent representation at the $l$-th layer of the teacher. Since the teacher utilizes both $x_i$ and $x_i^*$, we hypothesize that its $l$-th layer latent representation contains more informative features than the representation generated by the student, which relies solely on $x_i$. In our methodology, we incorporate this requirement into the training process of the student model by modifying the loss function as follows:
\begin{equation} \label{eq:loss_function}
L_s = (1 - \alpha) \cdot L(f_{student}(x, y)) + \alpha \cdot D(f_i^{(t)}, f_i^{(s)}).
\end{equation}
In this equation,  $L(f_{student}(x, y))$ represents the standard object detection loss, and $\alpha$ determines the relative influence of the teacher on the student’s learning process. It is important to emphasize that during the training phase, the student model leverages knowledge derived from $\{x^*_i\}_{i=1}^N$ by emulating the teacher’s latent representations, whereas during the testing phase, it relies exclusively on $x \in X$ to generate predictions.

\subsection{Implementation}
Given the object detection problem within the LUPI paradigm, the methodology for applying it to litter detection is as follows: Each object detection model uses both a teacher and a student network with identical layers, differing only at the input. The teacher receives a four-channel input-three-channel RGB plus a privileged information channel-while the student gets only the standard RGB input.

Selecting the privileged channel is challenging, especially for encoding bounding box information. Inspired by the Attention Spotlight principle in the human visual cortex \cite{spotlight}, a grayscale mask is generated for all bounding boxes, with each object class represented by a distinct shade. Preliminary tests showed this approach yielded the best results and was adopted as the privileged channel. Other forms, like saliency and depth prediction \cite{bartolo2024correlationobjectdetectionperformance}, did not show significant improvements.

Knowledge distillation from teacher to student occurs at the final backbone layer, where a feature representation vector is generated and Cosine Distance \cite{lab2wild} is used. This vector is incorporated into the student’s loss function, as defined in \eqref{eq:loss_function}.

The methodology was evaluated on five well-known object detection architectures-Faster R-CNN \cite{fasterrcnn}, RetinaNet \cite{retinanet}, FCOS \cite{fcos}, SSD \cite{ssd}, and SSDLite \cite{ssdlite}-across individual and multiple datasets. The approach, adaptable to any detection model, is shown in Figure \ref{fig:architecture}.

\subsection{Experimental Setup}\label{subsec:experiment}
To evaluate the methodology both within and across datasets, the publicly available SODA, BDW, and UAVVaste datasets were used. SODA was selected for training due to its varied-altitude images, offering practical, real-world data.

For preprocessing, SODA’s 829 images were tiled using a 3x3 grid (unlike the 5x5 in \cite{detect_litter}) based on hyper-parameter tuning, then resized to 1280x1280 pixels for high-resolution input. Privileged bounding box masks were generated on the tiled RGB images as grayscale masks. Min-Max normalization was applied to both RGB and mask images, standardizing pixel values to $[0,1]$. BDW and UAVVaste datasets were also resized to 1280x1280 pixels, but not tiled, as this was not part of their preprocessing. These datasets were used only for cross-dataset evaluation.

No data augmentation techniques were applied, as they were beyond the study’s scope. All detectors were trained with the Adam optimizer at a constant 0.0001 learning rate and no weight decay, based on preliminary tests showing Adam’s fast convergence and consistency. Early stopping with a patience of 8 was used to prevent overfitting, and all models were trained for 100 epochs. For post-processing, Non-Maximum Suppression (NMS) with an IoU threshold of 0.5 was applied to reduce background predictions.

\section{Results}

\subsection{Evaluation Metrics}
To evaluate the proposed methodology as outlined in Subsection \ref{subsec:experiment}, standard object detection metrics were adopted within the experimental framework. These included the COCO Detection metrics \cite{coco}, which follow the benchmark Mean Average Precision (mAP) at IoU thresholds of 0.5 and 0.75, as well as the averaged metric across a range from 0.5 to 0.95.
To complement the COCO metrics, three additional evaluation metrics were employed to facilitate a more thorough and nuanced assessment of the model's performance. Mean Precision evaluated how well the model identified correct detections whilst omitting false positives. Recall measured how completely the model detected all relevant objects. Finally, the F1 Score, calculated as the harmonic mean of Precision and Recall, served to evaluate each model’s performance by balancing its accuracy with its ability to detect all relevant objects.

\subsection{Performance Evaluation on the SODA Dataset}
Three experiments were conducted using the SODA dataset. The first involved training and evaluating the selected detectors on a 3 by 3 tiled version of the dataset, as detailed in Subsection \ref{subsec:experiment}, specifically for multi-label small litter detection. The second followed the same setup but assessed binary detection instead. The third focused on training and evaluating the detectors on a subset of images captured at an altitude of one meter, treating it as a binary litter detection task without tiling.

\begin{figure}[!htbp]
    \centering
    \includegraphics[width=0.47\textwidth]{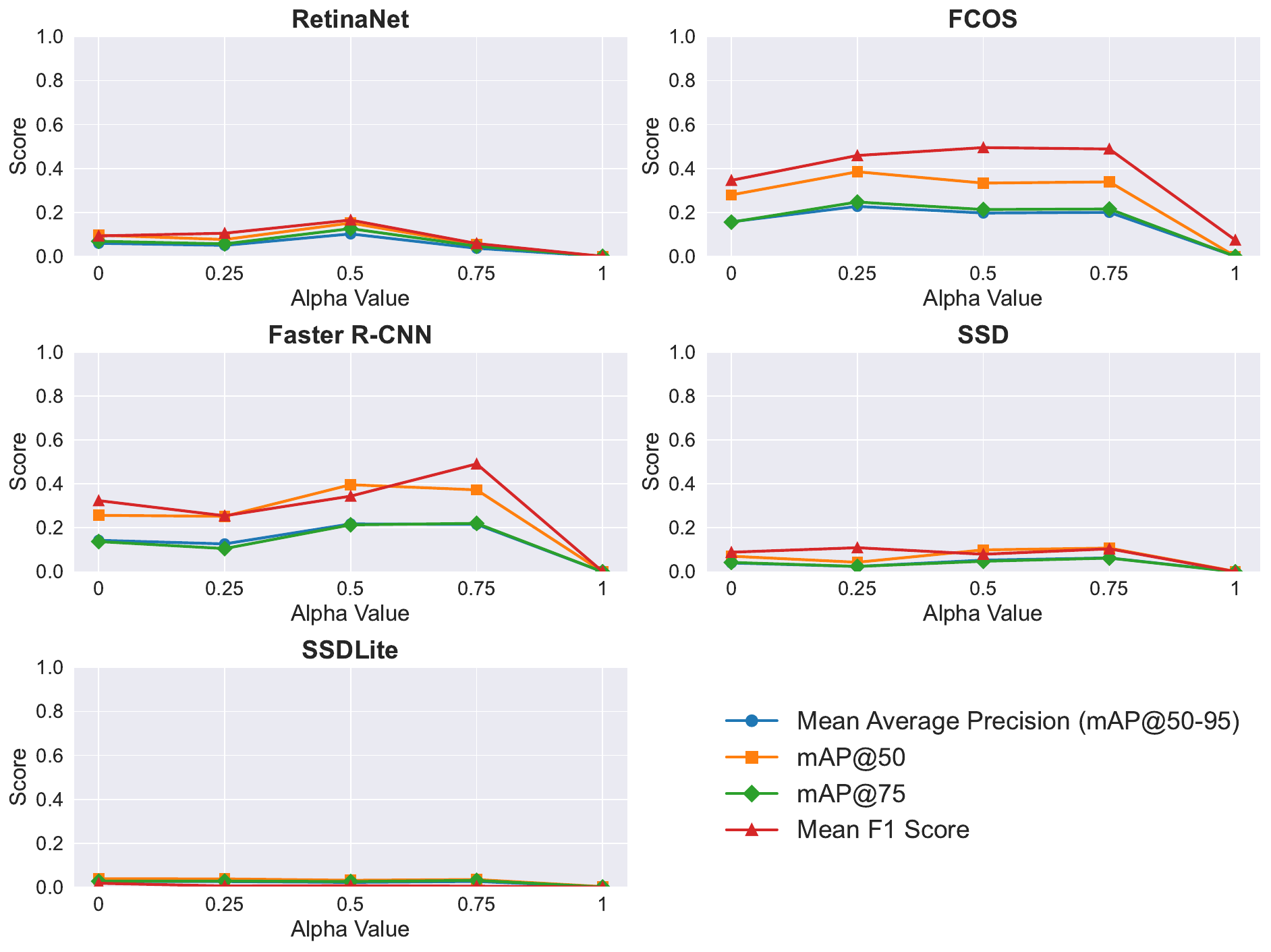}
    \caption{The Effect of the $\alpha$ Parameter on Student Model Performance (SODA Dataset - Tiled Multi-label Detection).}
    \label{fig:ablation}
\end{figure}

\begin{figure}[!htbp]
    \centering
    \includegraphics[width=0.47\textwidth]{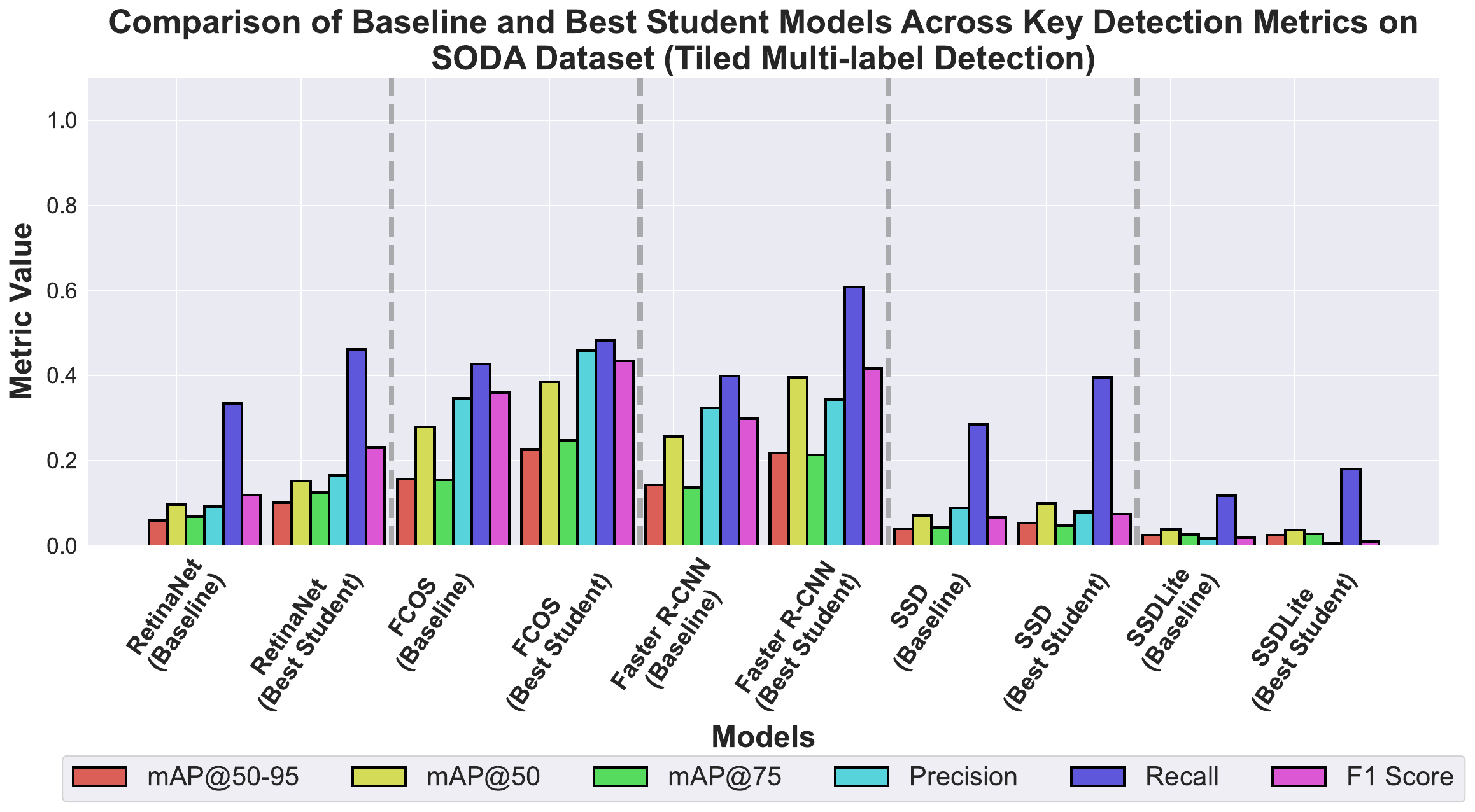}
    \caption{Comparison of Baseline and Best Student Models Across Key Detection Metrics on SODA Dataset (Tiled Multi-label Detection).}
    \label{fig:soda_multi}
\end{figure}

Although FCOS and RetinaNet have demonstrated superior results on the COCO benchmark—partly due to their more recent development and improved architectures—this pattern was also observed in the results of the first experiment for multi-label small litter detection. As shown in Figures \ref{fig:ablation}, and \ref{fig:soda_multi}, the results showed that FCOS emerged as one of the best performing detectors. Interestingly, Faster R-CNN outperformed RetinaNet in this specific task. Meanwhile, SSD and SSDLite yielded the lowest performance, yet all models still significantly benefited from the application of a teacher model, leading to notable improvements in detection accuracy.

An analysis was also carried out to investigate the impact of the teacher model on the performance of the student model, based on the influence parameter $\alpha$ as defined in \eqref{eq:loss_function}. For each of the selected models, student versions were trained using five different values for $\alpha: {0, 0.25, 0.5, 0.75, 1}$, as was done in \cite{lab2wild}.
As shown in Figure \ref{fig:ablation}, the $\alpha$ parameter had a noticeable effect on overall performance. On average, values between 0.25 and 0.5 resulted in higher mAP, while a value of 0.75 tended to yield better F1 Scores. It is also important to note that applying full teacher influence ($\alpha = 1$) frequently led to worse performance compared to omitting the teacher model altogether.

It is also worth highlighting that, when comparing the teacher models, the privileged information channel provided by the bounding box mask proved to be informative. This input enabled most models to more effectively learn the underlying target concept, as demonstrated in Table \ref{tab:teachers}. Interestingly, Faster R-CNN proved to be the most effective teacher model overall, demonstrating the greatest ability to grasp the true target concept, particularly in terms of small litter detection. However, FCOS and RetinaNet produced comparable results, suggesting that their architectures were also well suited to guiding student models. In contrast, SSD and SSDLite yielded weaker results as teacher models, which can be attributed in part to their simpler architecture. Nevertheless, these models still performed better than the baselines.

\begin{table}[ht]
\centering
\caption{Comparison of Teacher Models Across Key Detection Metrics on SODA Dataset (Tiled Multi-label Detection)}
\label{tab:teachers}
\resizebox{\columnwidth}{!}{
\begin{tabular}{|l|c|c|c|c|c|c|}
\hline
\textbf{Model} & \textbf{mAP@50-95} & \textbf{mAP@50} & \textbf{mAP@75} & \textbf{Precision} & \textbf{Recall} & \textbf{F1 Score} \\
\hline\hline
RetinaNet    & 0.88  & 0.92  & 0.91  & 0.76  & 0.97  & 0.85  \\\hline
FCOS         & 0.91  & 0.95  & 0.94  & 0.91  & 0.97  & 0.94  \\\hline
Faster R-CNN & \textbf{0.95}  & \textbf{0.99}  & \textbf{0.98}  & \textbf{0.96}  & \textbf{0.99}  & \textbf{0.97}  \\\hline
SSD          & 0.36  & 0.49  & 0.45  & 0.59  & 0.76  & 0.63  \\\hline
SSDLite      & 0.11  & 0.13  & 0.13  & 0.00  & 0.37  & 0.01  \\
\hline
\end{tabular}
}
\end{table}

\begin{figure}[!htbp]
    \centering
    \includegraphics[width=0.47\textwidth]{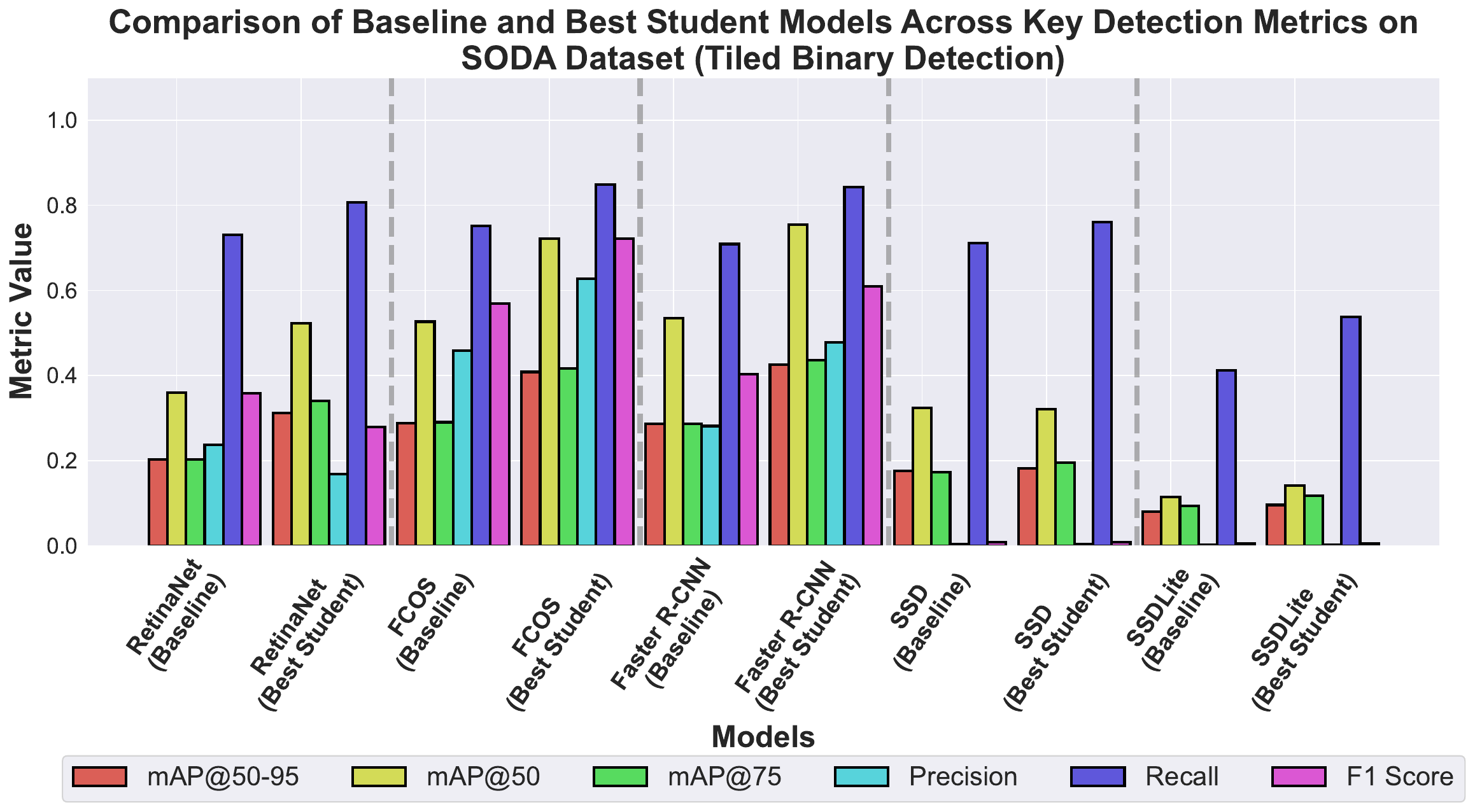}
    \caption{Comparison of Baseline and Best Student Models Across Key Detection Metrics on SODA Dataset (Tiled Binary Detection).}
    \label{fig:soda_single}
\end{figure}

\begin{figure}[!htbp]
    \centering
    \includegraphics[width=0.47\textwidth]{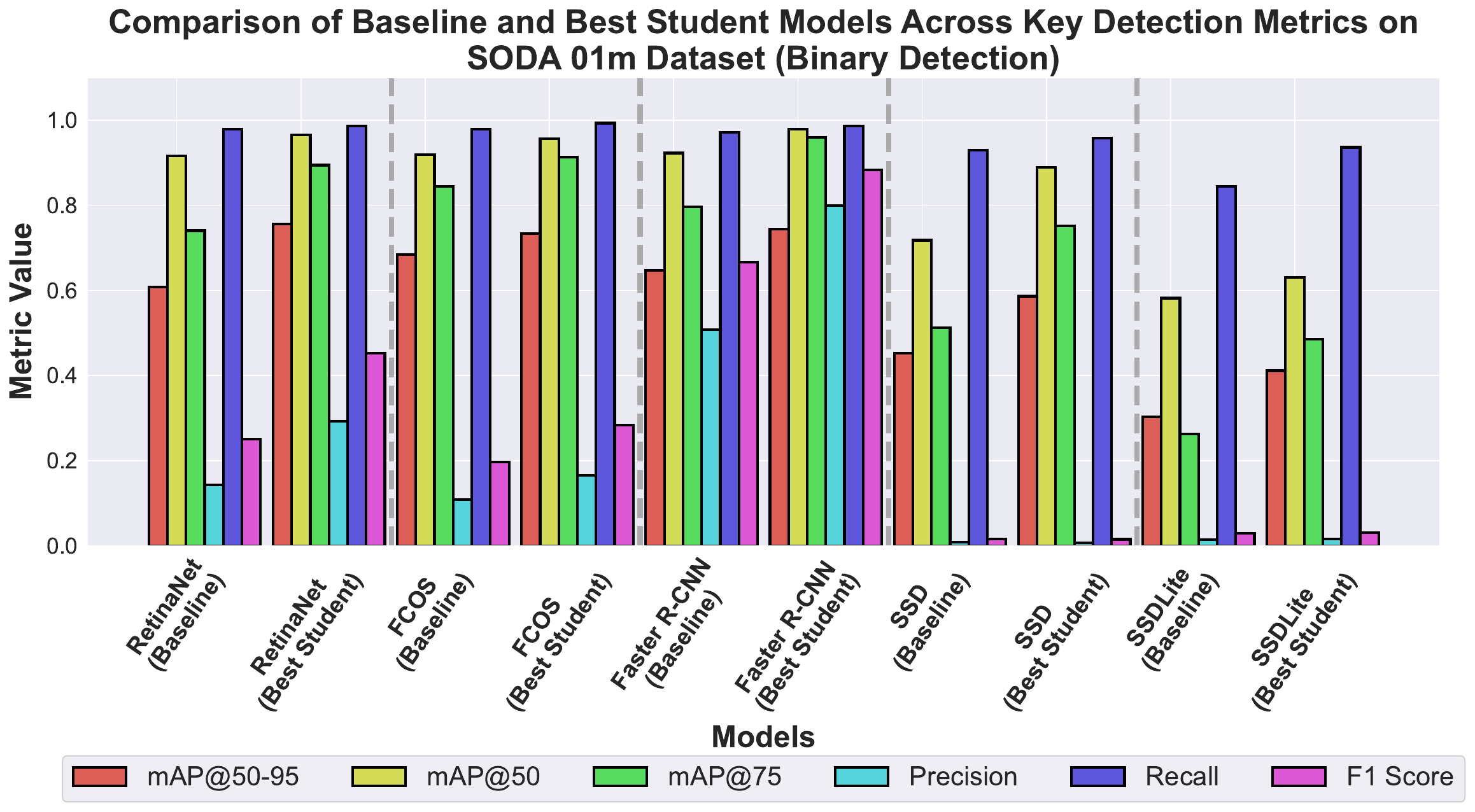}
    \caption{Comparison of Baseline and Best Student Models Across Key Detection Metrics on SODA 01m Dataset (Binary Detection).}
    \label{fig:soda_01m_single}
\end{figure}

Across all three experiments conducted on the SODA dataset, as illustrated in Figures \ref{fig:soda_multi}, \ref{fig:soda_single}, and \ref{fig:soda_01m_single}, there is a clear and consistent improvement when applying the proposed methodology to litter detection. This applies both to the localisation and classification components that define the detection task.
In the first experiment (Figure \ref{fig:soda_multi}), it was shown that applying the proposed methodology to address the problem of small litter detection, together with the use of tiling, led to a significant improvement in both mAP and F1 Score when comparing the performance of the student models to their respective baselines.

Similarly, in the second experiment (Figure \ref{fig:soda_single}), which focused on binary small litter detection, the methodology again demonstrated improved results compared to the baselines. While all models benefited from the approach, the improvements were more pronounced when comparing baseline models with their student counterparts. Models such as Faster R-CNN, FCOS, and RetinaNet exhibited notable improvements, whereas SSD and SSDLite achieved smaller improvements.

The third experiment aimed to assess whether the proposed methodology would still yield an improvement when applied to the task of close-range litter detection. At an altitude of one meter, the litter appears relatively large, effectively framing the task as a standard object detection problem. The results, as shown in Figure \ref{fig:soda_01m_single}, indicate a clear improvement, which in most cases is more pronounced than in the previous experiments. This suggests that the methodology remains proficient even when object scale is no longer a limiting factor.

\subsection{Cross-Dataset Performance Evaluation}

In addition to evaluating the trained models on the dataset used during training, two further experiments were carried out to assess how well the models would perform on external litter detection datasets. Specifically, the binary litter detection models were tested on the BDW and UAVVaste datasets, both of which also frame the problem as binary litter detection. Due to the characteristics of the BDW dataset, where bottle litter appears at a larger scale, the models trained on the SODA dataset at one meter altitude were used for inference. Conversely, the binary SODA tiled models were applied to the UAVVaste dataset, given its focus on small-scale litter.

\begin{figure}[!htbp]
    \centering
    \includegraphics[width=0.47\textwidth]{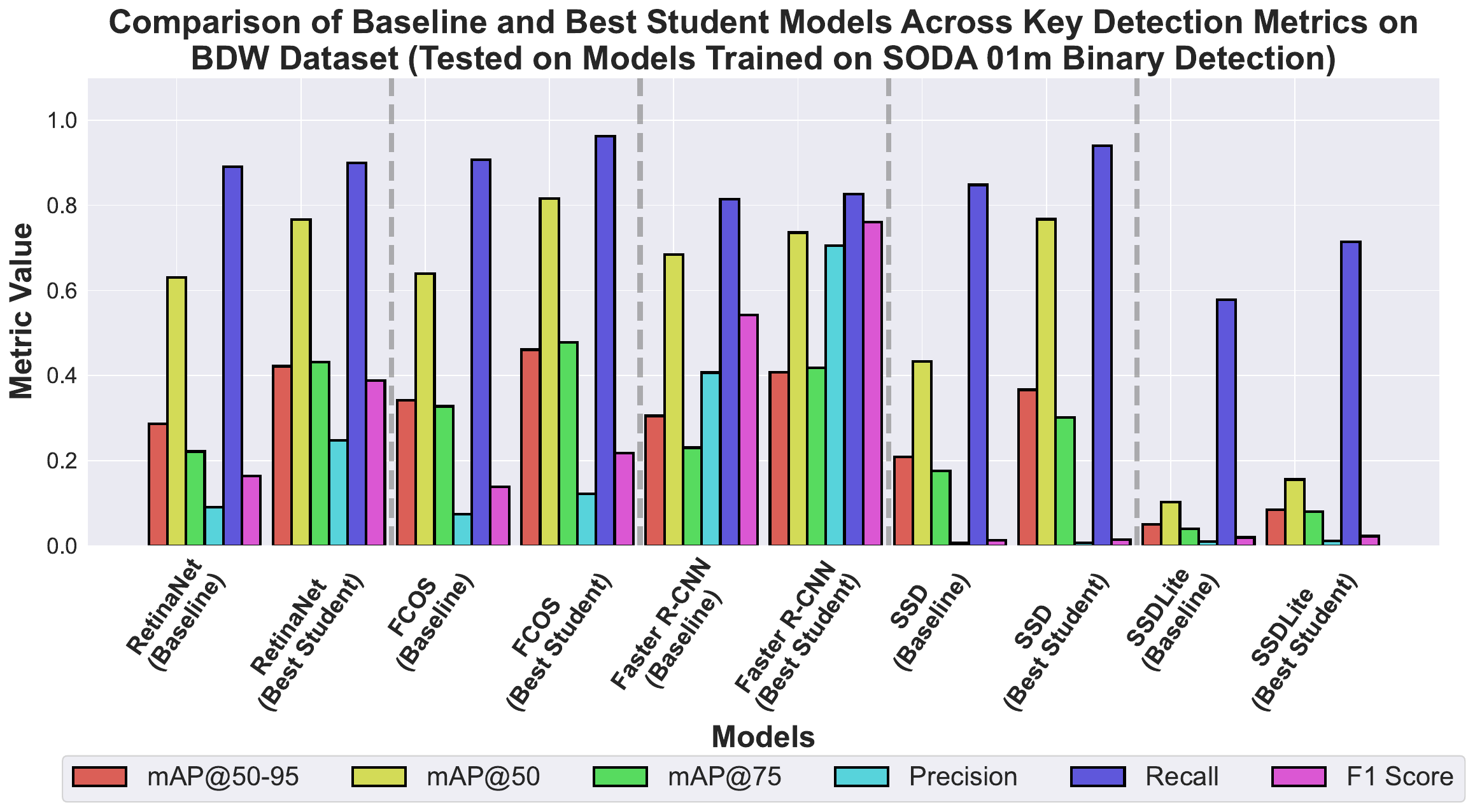}
    \caption{Comparison of Baseline and Best Student Models Across Key Detection Metrics on BDW Dataset (Tested on Models Trained on SODA 01m Binary Detection).}
    \label{fig:bdw}
\end{figure}

\begin{figure}[!htbp]
    \centering
    \includegraphics[width=0.47\textwidth]{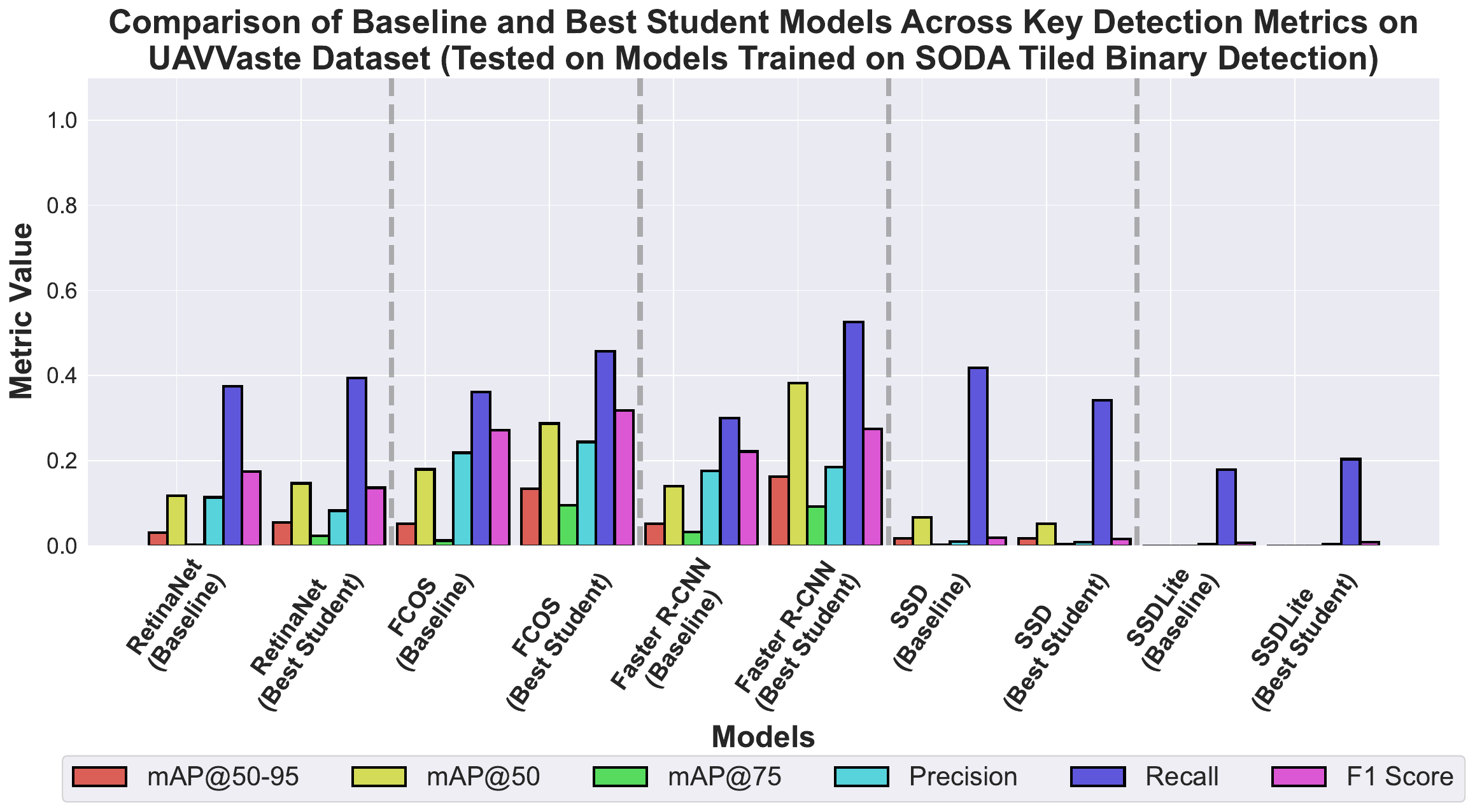}
    \caption{Comparison of Baseline and Best Student Models Across Key Detection Metrics on UAVVaste Dataset (Tested on Models Trained on SODA Tiled Binary Detection).}
    \label{fig:uavvaste}
\end{figure}

In both cases (Figures \ref{fig:bdw}, and \ref{fig:uavvaste}), the student models continued to outperform their corresponding baselines, demonstrating that the benefits of the proposed methodology extend beyond the original training data. While SSD and SSDLite followed a similar trend to previous experiments, showing only marginal gains, the overall advantage of adopting the proposed approach remains evident.

The results of these experiments demonstrate substantial improvements across five object detection models applied to litter detection, with consistent advancements observed throughout. Notably, no architectural changes were made between the baseline and student models, nor was there any increase in parameters or layers. Nevertheless, performance improved, albeit with slightly longer training times due to the added cost of generating privileged information and training a teacher model. As each result reflects a single experimental run, statistical analysis was not applicable.

\section{Conclusion}

This study proposed a novel methodology that integrates privileged information and knowledge distillation to improve litter detection, all without increasing model parameters or affecting inference time. The methodology was tested across five widely used object detectors, addressing different detection challenges, including small litter detection and standard object detection for objects at varying scales. A key contribution of this work is the introduction of a novel technique for encoding bounding box information, which is fed to the model as a binary mask. This approach was found to be informative, aiding the model in guiding the detection process more effectively.
The results demonstrated consistent improvements when applying this methodology, both within the models trained on the SODA dataset and through cross-dataset evaluations on the BDW and UAVVaste litter detection datasets. These findings illustrate that the proposed methodology not only boosts performance on the dataset it was trained on, but also generalises well to other litter detection datasets. Importantly, these improvements were achieved without the need to increase model complexity or add new layers, making the approach both efficient and practical. As a natural extension of this work, future experiments could explore the generalisation capability of the approach on broader and more diverse benchmarks, including Pascal VOC and COCO.

\bibliographystyle{IEEEtran}
\bibliography{references}

\begin{thebibliography}{10}
\providecommand{\url}[1]{#1}
\csname url@samestyle\endcsname
\providecommand{\newblock}{\relax}
\providecommand{\bibinfo}[2]{#2}
\providecommand{\BIBentrySTDinterwordspacing}{\spaceskip=0pt\relax}
\providecommand{\BIBentryALTinterwordstretchfactor}{4}
\providecommand{\BIBentryALTinterwordspacing}{\spaceskip=\fontdimen2\font plus
\BIBentryALTinterwordstretchfactor\fontdimen3\font minus \fontdimen4\font\relax}
\providecommand{\BIBforeignlanguage}[2]{{%
\expandafter\ifx\csname l@#1\endcsname\relax
\typeout{** WARNING: IEEEtran.bst: No hyphenation pattern has been}%
\typeout{** loaded for the language `#1'. Using the pattern for}%
\typeout{** the default language instead.}%
\else
\language=\csname l@#1\endcsname
\fi
#2}}
\providecommand{\BIBdecl}{\relax}
\BIBdecl

\bibitem{kaza2018waste}
S.~Kaza, L.~C. Yao, P.~Bhada-Tata, and F.~V. Woerden, \emph{What a Waste 2.0: A Global Snapshot of Solid Waste Management to 2050}.\hskip 1em plus 0.5em minus 0.4em\relax Washington, DC: World Bank, 2018.

\bibitem{zerowaste}
D.~Bashkirova, M.~Abdelfattah, Z.~Zhu, J.~Akl, F.~Alladkani, P.~Hu, V.~Ablavsky, B.~Calli, S.~A. Bargal, and K.~Saenko, ``Zerowaste dataset: Towards deformable object segmentation in cluttered scenes,'' in \emph{Proceedings of the IEEE/CVF Conference on Computer Vision and Pattern Recognition (CVPR)}, 6 2022, pp. 21\,147--21\,157.

\bibitem{taco2020}
P.~F. Proença and P.~Simões, ``Taco: Trash annotations in context for litter detection,'' \emph{arXiv preprint arXiv:2003.06975}, 2020.

\bibitem{uavvaste}
M.~Kraft, M.~Piechocki, B.~Ptak, and K.~Walas, ``Autonomous, onboard vision-based trash and litter detection in low altitude aerial images collected by an unmanned aerial vehicle,'' \emph{Remote Sensing}, vol.~13, no.~5, 2021.

\bibitem{soda_dataset}
D.~Pisani, D.~Seychell, C.~J. Debono, and M.~Schembri, ``Soda: A dataset for small object detection in uav captured imagery,'' in \emph{2024 IEEE International Conference on Image Processing (ICIP)}, 2024, pp. 151--157.

\bibitem{bottle_detection}
J.~Wang, W.~Guo, T.~Pan, H.~Yu, L.~Duan, and W.~Yang, ``Bottle detection in the wild using low-altitude unmanned aerial vehicles,'' in \emph{2018 21st International Conference on Information Fusion (FUSION)}, 2018, pp. 439--444.

\bibitem{mju_waste}
T.~Wang, Y.~Cai, L.~Liang, and D.~Ye, ``A multi-level approach to waste object segmentation,'' \emph{CoRR}, vol. abs/2007.04259, 2020.

\bibitem{plastopol}
M.~Córdova, A.~Pinto, C.~C. Hellevik, S.~A.-A. Alaliyat, I.~A. Hameed, H.~Pedrini, and R.~d.~S. Torres, ``Litter detection with deep learning: A comparative study,'' \emph{Sensors}, vol.~22, no.~2, 2022.

\bibitem{detect_litter}
D.~Pisani and D.~Seychell, ``Detecting litter from aerial imagery using the soda dataset,'' in \emph{2024 IEEE 22nd Mediterranean Electrotechnical Conference (MELECON)}, 2024, pp. 897--902.

\bibitem{yolo}
J.~Redmon, S.~Divvala, R.~Girshick, and A.~Farhadi, ``You only look once: Unified, real-time object detection,'' 2016, cite arxiv:1506.02640.

\bibitem{fasterrcnn}
S.~Ren, K.~He, R.~B. Girshick, and J.~Sun, ``Faster {R-CNN:} towards real-time object detection with region proposal networks,'' \emph{CoRR}, vol. abs/1506.01497, 2015.

\bibitem{ssd}
W.~Liu, D.~Anguelov, D.~Erhan, C.~Szegedy, S.~E. Reed, C.~Fu, and A.~C. Berg, ``{SSD:} single shot multibox detector,'' in \emph{Computer Vision - {ECCV} 2016 - 14th European Conference, Amsterdam, The Netherlands, October 11-14, 2016, Proceedings, Part {I}}, ser. Lecture Notes in Computer Science, B.~Leibe, J.~Matas, N.~Sebe, and M.~Welling, Eds.\hskip 1em plus 0.5em minus 0.4em\relax Springer, 2016, vol. 9905, pp. 21--37.

\bibitem{retinanet}
T.~Lin, P.~Goyal, R.~B. Girshick, K.~He, and P.~Doll{\'{a}}r, ``Focal loss for dense object detection,'' in \emph{2017 {IEEE} International Conference on Computer Vision ({ICCV})}.\hskip 1em plus 0.5em minus 0.4em\relax Venice, Italy: {IEEE} Computer Society, 2017, pp. 2999--3007.

\bibitem{lupi}
V.~Vapnik and A.~Vashist, ``A new learning paradigm: Learning using privileged information,'' \emph{Neural networks : the official journal of the International Neural Network Society}, vol.~22, pp. 544--57, 07 2009.

\bibitem{Vapnik2015LearningUP}
V.~N. Vapnik and R.~Izmailov, ``Learning using privileged information: similarity control and knowledge transfer,'' \emph{J. Mach. Learn. Res.}, vol.~16, pp. 2023--2049, 2015.

\bibitem{learning2rank}
V.~Sharmanska, N.~Quadrianto, and C.~H. Lampert, ``Learning to rank using privileged information,'' in \emph{2013 IEEE International Conference on Computer Vision}, 2013, pp. 825--832.

\bibitem{lupi_classification}
S.~Wang, S.~Chen, T.~Chen, and X.~Shi, ``Learning with privileged information for multi-label classification,'' \emph{Pattern Recognition}, vol.~81, pp. 60--70, 2018.

\bibitem{distillation1}
G.~Habib, T.~jan Saleem, S.~M. Kaleem, T.~Rouf, and B.~Lall, ``A comprehensive review of knowledge distillation in computer vision,'' 2024.

\bibitem{distillation2}
Z.~Zheng, R.~Ye, Q.~Hou, D.~Ren, P.~Wang, W.~Zuo, and M.-M. Cheng, ``Localization distillation for object detection,'' 2022.

\bibitem{spotlight}
S.~A. McMains and D.~C. Somers, ``Multiple spotlights of attentional selection in human visual cortex,'' \emph{Neuron}, vol.~42, no.~4, pp. 677--686, 2004.

\bibitem{bartolo2024correlationobjectdetectionperformance}
M.~Bartolo and D.~Seychell, ``Correlation of object detection performance with visual saliency and depth estimation,'' 2024.

\bibitem{lab2wild}
K.~Makantasis, K.~Pinitas, A.~Liapis, and G.~N. Yannakakis, ``From the lab to the wild: Affect modeling via privileged information,'' \emph{IEEE Transactions on Affective Computing}, vol.~15, no.~2, pp. 380--392, 2024.

\bibitem{fcos}
Z.~Tian, C.~Shen, H.~Chen, and T.~He, ``{FCOS: Fully Convolutional One-Stage Object Detection},'' in \emph{2019 {IEEE/CVF} International Conference on Computer Vision ({ICCV})}.\hskip 1em plus 0.5em minus 0.4em\relax Seoul, Korea (South): {IEEE}, 2019, pp. 9626--9635.

\bibitem{ssdlite}
M.~Sandler, A.~Howard, M.~Zhu, A.~Zhmoginov, and L.-C. Chen, ``Mobilenetv2: Inverted residuals and linear bottlenecks,'' 2019.

\bibitem{coco}
T.-Y. Lin, M.~Maire, S.~J. Belongie, J.~Hays, P.~Perona, D.~Ramanan, P.~Doll{\'a}r, and C.~L. Zitnick, ``Microsoft {COCO:} common objects in context,'' in \emph{Computer Vision - {ECCV} 2014 - 13th European Conference, September 6-12, 2014}, ser. Lecture Notes in Computer Science, D.~J. Fleet, T.~Pajdla, B.~Schiele, and T.~Tuytelaars, Eds., vol. 8693.\hskip 1em plus 0.5em minus 0.4em\relax Cham, Switzerland: Springer, 2014, pp. 740--755.

\end{thebibliography}

\end{document}